\documentclass{article}
\usepackage[utf8]{inputenc}
\usepackage{iclr2016_conference,times}
\usepackage{hyperref}
\usepackage{url}
\usepackage{times}
\usepackage{epsfig}
\usepackage{graphicx}
\usepackage{amsmath}
\usepackage{amssymb}

\title{Fine-grained pose prediction,\\ normalization, and recognition}

\author{Ning Zhang \\
Snapchat Inc.\\
Venice, CA, USA \\
\texttt{\{ning.zhang\}@snapchat.com} \\
\And
Evan Shelhamer ~~ Yang Gao ~~ Trevor Darrell \\
Department of Computer Science \\
University of California, Berkeley \\
Berkeley, CA, USA \\
\texttt{\{shelhamer, yg, trevor\}@eecs.berkeley.edu} \\
}

\begin{document}
\newcommand{\todo}[1]{{\color{red} {\bf TODO:} \it #1}}
\newcommand{\ning}[1]{{\color{blue}{\bf Ning:} \it #1}}
\newcommand{\evan}[1]{{\color{green}{\bf Evan:} \it #1}}
\maketitle
\begin{abstract}
Pose variation and subtle differences in appearance are key challenges to fine-grained classification.
While deep networks have markedly improved general recognition, many approaches to fine-grained recognition rely on anchoring networks to parts for better accuracy.
Identifying parts to find correspondence discounts pose variation so that features can be tuned to appearance.
To this end previous methods have examined how to find parts and extract pose-normalized features.
These methods have generally separated fine-grained recognition into stages which first  localize parts using hand-engineered and coarsely-localized proposal features, and then separately learn deep descriptors centered on inferred part positions.
We unify these steps in an end-to-end trainable network supervised by keypoint locations and class labels that localizes parts by a
fully convolutional network to focus the learning of feature representations for the fine-grained classification task.
Experiments on the popular CUB200 dataset show that our method is state-of-the-art and suggest a continuing role for strong supervision.
\end{abstract}

\section{Introduction}

Fine-grained classification is a challenging task due to striking variations in pose, subtle differences in appearance between classes, and strong intra-class variety. Fine-grained classification has made progress in improving accuracy and 
weakening supervision and now covers several domains including aircraft \citep{maji13fine-grained}, cars \citep{StarkKrauseBMVC12}, plants \citep{AngelovaCVPR13,Anelia13,leafsnap,nilsback_visual_2006,nilsback_automated_2008,VantageFramesCVPR13}, and animal breeds \citep{Liu_Dogs_2012}.

Unlike basic-level object classification, where learning holistic representations alone continues to improve recognition, fine-grained classification often relies on correspondence to align part representations so that the subtle distinguishing features of different classes can be learned.
To identify these correspondences, previous work \citep{eccv14, BransonHBP14} has focused on using strong supervisions for part localizations and feature transformations that can discount the nuisance variables of pose and viewpoint.
While recent work \citep{transform_net, bilinear:arxiv, nopart:cvpr15} has made impressive progress without strong supervision, we show that the correspondence gained from end-to-end keypoint and classification training improves fine-grained recognition accuracy.


Deep learning has made dramatic progress on not only image classification, but detection \citep{overfeat, rcnn}, keypoint prediction \citep{TompsonJLB14}, and semantic segmentation \citep{fcn}.
These tasks require not only recognition but localization.
Whether or not a deep network can effectively learn all necessary invariances to pose is an open question.
We believe it is still critical to model parts and pool features into a pose-independent representation to best distinguish between similar subordinate classes.
This makes good use of limited training data by reserving model capacity for appearance given the discounting of pose.


Real-world fine-grained recognition cannot rely on side-channel information like knowledge of the true bounding boxes or parts. We are inspired by the object and part detection approach of  \citep{eccv14}, which makes use of the state-of-the-art R-CNN \citep{rcnn} detection method, and learns a deep representation of part-level features. However, the part recall of this model is limited by proposals based on hand-engineered features and the separate optimization of part localization and deep feature learning stages.

In this paper, we propose an end-to-end trainable deep network to simultaneously localize parts using a spatially fine-grained detection model, form descriptors over inferred part locations, and classify a categorically fine-grained label space.
The network goes directly from pixels to parts via fully convolutional keypoint localization network.
We design a semantic pooling layer---which we call the coordinate transfer layer---to pool feature maps using predicted keypoints.
The fine-grained classification is then decided from the combined representations.
Classification errors are back-propagated to the selected parts to drive the learning of discriminative, pose-normalized features.
To our knowledge, ours is the first  approach to incorporate pose prediction and normalization in a unified  deep fine-grained model. Our end-to-end model achieves state-of-the-art results on the standard Caltech-UCSD bird benchmark \citep{DatasetCUB200} for
fine-grained recognition.


\begin{figure*}[t]
\begin{center}
   \includegraphics[width=1.0\linewidth]{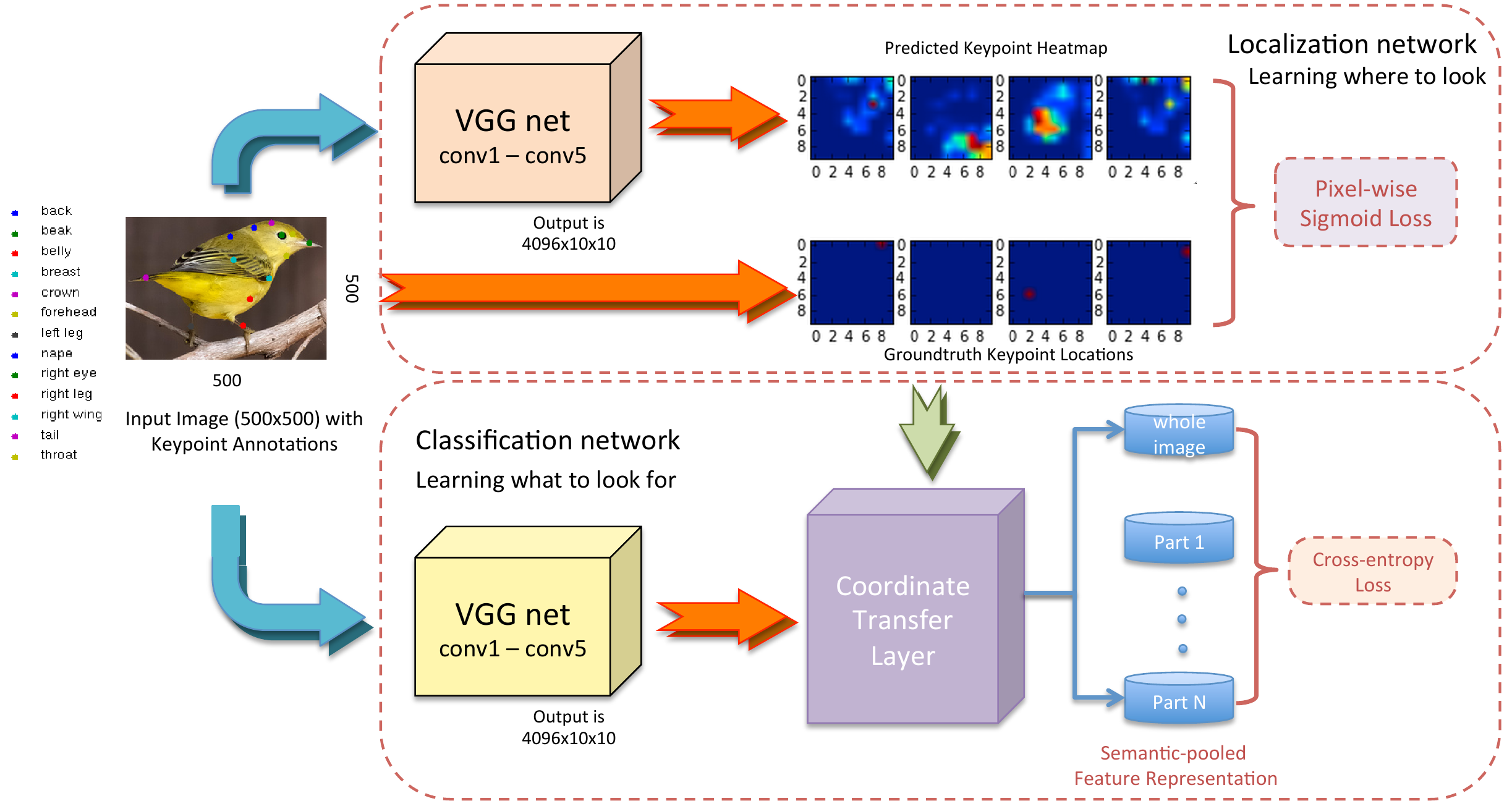}
\end{center}
   \caption{\textbf{Overview of our model.}
  The network consists of two main modules: 1) a localization network for learning where to look
and 2) a classification network which uses a coordinate transfer layer to semantically pool part features; these are jointly used to learn the fine-grained classifier.
 } 
\label{fig:concept}
\end{figure*}

\section{Related work}

Our approach draws on the significant literature on fine-grained classification methods, and on recent successes of deep nets for image classification \citep{krizhevsky, vgg} and transfer learning \citep{decaf} especially their fully convolutional realizations; both areas are reviewed below in turn.

\subsection{Fine-grained classification}
Several approaches are based on segmentation methods  \citep{Tricos_Chai_ECCV12,iccv13_alignment,iccv13_partmatching, iccv13_symbiotic}  and others are based on detecting and extracting features from certain parts of objects, such as ~\citep{BirdletsFarrellICCV11, Liu_Dogs_2012, poof, iccv13_keypoint, Goering14:NPT}. 
\citep{eccv14} adapt the R-CNN detector for not only whole object but part localization and recognition.
This brought fine-grained classification into the more realistic setting of not having oracle knowledge of the bounding box at test time.
However, this approach still relies on a proposal pipeline.
Parts pose a problem by their small size and potentially subtle boundaries that leads to poor proposal recall and limited the pose normalization of the part R-CNN to head and body.
The proposed model is able to detect and pose normalize across all keypoints through fully convolutional search.
 \citep{BransonHBP14} analyze the space of staged detection, alignment, and classification pipelines to reveal the role of feature learning in fine-grained recognition.
The choice of learned features makes the most difference but pose has its own contribution.
This reinforces the need for the end-to-end learning and pose integration that our work advances.
\citep{deepLAC:cvpr15} formulate a joint localization, alignment, and classification model for fine-grained classification.
Subnetworks regress part bounding boxes, match to clustered part templates, and classify.
This approach improves on separate models but is restricted to head and body keypoints and coarse localization by bounding box regression.
Although localization and classification are jointly tuned, the ``valve link function'' for pose normalization is fixed and nonparametric.
In contrast our proposed network is capable of fine-grained localization by fully convolutionally predicting all keypoint types and the entire model is learned jointly.

 \citep{nopart:cvpr15} achieve fine-grained recognition without part annotations.
Inferring latent parts by co-segmentation drives pose alignment for learning R-CNN detectors of the whole object and the discovered parts.
Parts are mined from co-segmentation by a diversity clustering that is not guaranteed to be discriminative, unlike the task-driven end-to-end training of our model.
While the latent parts are learned in disjoint stages, weakening supervision is an exciting direction for fine-grained recognition.
Such weak signals could be incorporated into our fully convolutional training analogously to the bounding box mining of \citep{dai2015boxsup} for segmentation masks. The spatial transformer network of \citep{transform_net} strives for spatial invariance by end-to-end learning transformations and is shown to be effective for fine-grained categorization.

Our method is also related to attention models. Olshausen et al. proposed a neural model of visual attention which is based on dynamic gating of information flow at stage of representation within a deep network \citep{olshausen:neurobiological}. Recent attention models \citep{mnih_attention, ba_attention} use sequential deep learning models, such as recurrent neural network or LSTM to learn glimpse location based on last glimpse. In \citep{sermanet_attention}, the authors use attention models to learn where to look for as glimpses for fine-grained classification. Compared with attention models, our method is not restricted to sequential and can directly exploit available strong supervision. While appealing in generality, these attention-based approaches do not exceed performance of recent direct vision approaches.

\subsection{Fully convolutional networks}

The fully convolutional networks (FCNs) of \citep{fcn} are designed for pixel-wise prediction.
Every layer in an FCN computes a local operation on relative spatial coordinates.
There are no fully connected layers to restrict dimensions.
In this way, an FCN can take an input of any size and produce an output of corresponding dimensions.

Inference and learning are done on whole images at a time.
The dense forward pass and backpropagation are highly efficient since intermediate features and gradients are effectively cached.
This makes it possible to take advantage of the complete ground truth in training pixel-to-pixel tasks such as semantic segmentation or in our case keypoint prediction.

\citep{TompsonJLB14} learn an end-to-end fully convolutional network for human pose. However, this model predicts keypoints alone and does not explore their role in recognition. In this work, the keypoints guide the fine-grained classification model by pooling deep features into a pose-normalized form that is learned end-to-end.

\citep{shubham} combine convolutional score maps with inferred viewpoint priors to regress to keypoint locations. Their viewpoint and keypoint prediction depend on detections from R-CNN. Joint learning is prevented by the independent proposal and bounding box recognition stages.
The fully convolutional keypoints to fine-grained classification network proposed in this work does not rely on separate detections or bounding box knowledge.

\begin{figure}[t]
\begin{center}
   \includegraphics[width=0.5\linewidth]{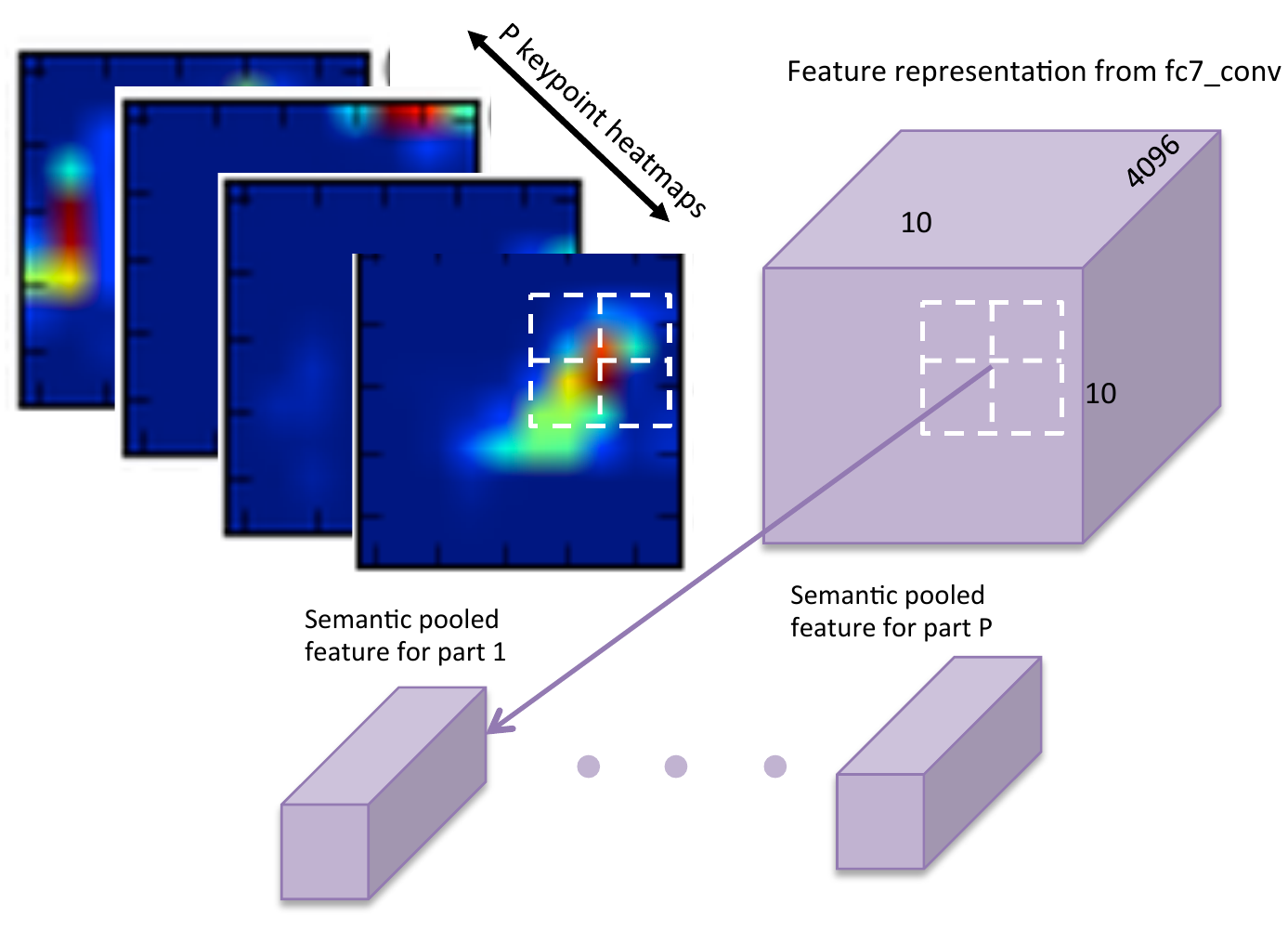}
\end{center}
   \caption{  
   The coordinate transfer layer takes two inputs: the keypoint heatmap and the feature representation activations. For each part, the layer pools the feature based on the small surrounding neighborhood around the argmax point of the heatmap, as shown in white rectangle. For $P$ different parts, the layer pools $P$ semantic features and stacks them together as the output.}
\label{fig:coordinatetransfer}
\end{figure}

\section{Fully convolutional part model for fine-grained categorization} 
The overview architecture of our model is shown in Figure \ref{fig:concept}. The goal is to define a unified deep network to simultaneously learn to localize parts and semantically pool the feature representations accordingly. The network consists of two main modules: 1) localization net learning where to look and 2) classification net which has semantic pooling to learn the fine-grained classifier.

We initialize our network from the VGG-16 network \citep{vgg} as pretraining. With its further parameters and layers of nonlinearity, the VGG net was a winner of the ILSVRC 2014 challenge. It has since proven useful through transfer learning to other vision tasks \citep{hypercolumn, fcn} like semantic segmentation that require both recognition and localization. We use the publicly available pretained VGG-16 weights and implement our model with the Caffe \citep{caffe} framework.

\subsection{Keypoint prediction}
First, we cast the VGG network into fully convolutional form following \cite{fcn}.
This equips the network for dense feature extraction on any input image size.
Although the original input size of VGG net is $224 \times 224$ pixels, the last layer receptive field size is actually $404 \times 404$. 
This is the size of the spatial context for each keypoint prediction.
For finer part resolution, we upsample the input image (which we note can be composed with the filters of the first layer) to $500 \times 500$ for an ultimate feature map of dimensions $10 \times 10 \times 4096$.
We score keypoints by a $1 \times 1$ convolution layer.

The keypoint location scoring is learned through end-to-end training from a pixelwise sigmoid cross-entropy loss function.
The ground truth keypoint locations are coded as a binary map over output locations.
The output centered at closest point to the true keypoint is coded as a positive while all other outputs are negatives.
A keypoint may not be visible in all inputs; these missing keypoints are coded as negatives at every location.

The pixelwise sigmoid cross entropy loss function is
\[
l = \frac{1}{N} \sum_{n=1}^N \sum_{i=1}^W \sum_{j=1}^H [p_{ij}^n log\hat{p}_{ij}^n + (1-p_{ij}^n)log(1-\hat{p}_{ij}^n)]
\]
where $p_{ij}^n$ is the ground truth keypoint map of $n$th example at location $(i,j)$ and $\hat{p}_{ij}^n$ is the sigmoid output at location $(i,j)$ of $n$th example.

\subsection{Coordinate transfer layer}

It is not enough to have keypoint locations:
rich, part-tuned features are needed to capture fine-grained differences.
In order to direct feature learning to this purpose, we pool features semantically by keypoint identity.
This is carried out by our novel coordinate transfer layer.
It takes two inputs: the keypoint location map and the feature activations. 
In the forward pass, the layer pools the features maps in a focused, surrounding neighborhood of the part, as shown by the white rectangle in Figure \ref{fig:coordinatetransfer} and then stack the pooled features together as the pose-normalized part representation. 
During back-propagation, the layer propagates the classification error to the corresponding part region in the lower feature representation. During training the ground truth part locations are pooled whereas the predicted keypoint locations are pooled at test time.



\subsection{Joint representation}
For fine-grained class recognition, we concatenate the pose-normalized part feature and the holistic representation of the whole image together and add a fully connected layer  on top of the fused representation.
The classification training is done by the softmax loss.
At testing neither the bounding box or part annotations are given; everything is inferred from the image. For the holistic representation, we also optionally employ an additional layer implementing the recent bilinear feature space of \citep{bilinear:arxiv} and its compact extension \citep{compact_bilinear}.



\subsection{Training}
The whole network can be trained end-to-end for joint learning of all model parameters.
The architecture is a directed acyclic graph and the coordinate transfer layer gradient back-propagates fine-grained recognition error while respecting localization.

Given the size of the network, we adopt a staged approach to learning before end-to-end fine-tuning.
This ensures reasonable accuracy of the keypoint and classification tasks before complete feature learning to avoid divergence from the pre-training.

Our training proceeds as follows:
\begin{enumerate}
	\item Keypoint localization is fine-tuned from the VGG net for fixed features and random initialization of the last score layer.
	\item Once converged, we fine-tune the localization network at a smaller learning rate.
	\item Once localization network has converged, the classifiers are trained for fixed semantic pooling
	on ground truth part annotations.
	\item Once converged, we fine-tune all layers at a smaller learning rate.
\end{enumerate}



\subsection{Compact bilinear representation}
 \citep{bilinear:arxiv} have shown impressive results on fine-grained classification by using bilinear pooling to encode the deep representation. However, those features are high dimensional (\~262k) and impractical for spatial tasks like keypoints or segmentation. \citep{compact_bilinear} propose compact representations with same discriminative power which allow back-propagation and end-to-end training. We use the "Random Maclaurin Projection" of \citep{compact_bilinear} to yield a deep representation that of only 5k dimensions. 

\subsection{Finetuned part nets}
The coordinate transfer layer provides a way to use the feature representation around each keypoint and unify the whole pose-normalized representation into one network. However, one disadvantage is that it lacks context due to the smaller corresponding receptive windows. An alternative way of utilizing keypoints for pose-normalized representation is to crop part images as \citep{eccv14}. Following \citep{eccv14}, we crop two parts: head and body using the predicted keypoints and finetuned part networks using the crop part images.

\section{Experimental results}
In this section, we present a comparative performance evaluation of our proposed method.
Specifically, we conduct experiments on the widely-used fine-grained benchmark Caltech-UCSD bird dataset \citep{DatasetCUB200} (CUB200-2011).
The classification task is to discriminate among 200 species of birds, and is challenging for computer vision systems due to the high degree of similarity between categories.
It contains 11,788 images of 200 bird species. Each image is annotated with its bounding box and the image coordinates of fifteen keypoints: the beak, back, breast, belly, forehead, crown, left eye, left leg, left wing, right eye, right leg, right wing, tail, nape and throat. We train and test on the splits included with the dataset, which contain around 30 training samples for each species.

\begin{table}[t]
\centering
\begin{tabular}{|l|c|c|c|c|c|}
\hline
Method  & \multicolumn{2}{c|}{Train phase} & \multicolumn{2}{c|}{Test phase}    & Accuracy\\
\hline
& BBox & Parts & BBox & Parts & \\ \hline
DPD+DeCAF feature ~\citep{decaf} & \checkmark & \checkmark & \checkmark & & 64.96\% \\
POOF~\citep{poof} & \checkmark & \checkmark & \checkmark & & 56.78\% \\ 
Symbiotic Segmentation~\citep{iccv13_symbiotic} & \checkmark & & \checkmark & \ &  59.40\% \\
Alignment~\citep{iccv13_alignment} &  \checkmark & & \checkmark & & 62.70\%\\
Part-based RCNN with BBox\citep{eccv14} & \checkmark & \checkmark & \checkmark & &  76.37\% \\
Part-based RCNN\citep{eccv14} & \checkmark & \checkmark &  &  &73.89\% \\
Pose normalized CNNs \citep{BransonHBP14}& \checkmark & \checkmark &   & & 75.70\% \\
Co-segmentation \citep{nopart:cvpr15} & \checkmark &  &  & & 82.0\%\\ 
Two-level attention \citep{attention:cvpr15} & & & & & 69.7\%\\
Bilinear model \citep{bilinear:arxiv} & & & & & 84.0\% \\
Spatial transformer networks \citep{transform_net} & & & & & 84.1\% \\
\hline
fc7 feature of VGG on whole image &  & & & & 58.34\%\\
fc7 feature of VGG-ft on whole image & & & &  & 71.73\% \\
\hline
Keypoint features baseline& & \checkmark &&& 65.10\% \\ 
Our model with fc7 feature &  & \checkmark & & & 75.04\% \\ 
Our model with compact bilinear feature &   & \checkmark &  & & 83.00\% \\
Our model with compact bilinear feature on ft-part nets &  & \checkmark &&&\textbf{85.92\%} \\
\hline
\end{tabular}
\caption{Comparison with other methods on fine-grained Classification results.} 
\label{table:finegrainedres}
\end{table}


\begin{table}[t]
\centering	
\begin{tabular}{|l|r|r|r|r|}
\hline
Parts & $\alpha=0.02$ & $\alpha=0.05$ & $\alpha=0.08$ & $\alpha=0.10$\\
\hline
Back&    9.4\%&   46.8\%&   74.8\%&   85.6\% \\
Beak&   12.7\%&   62.5\%&   89.1\%&   94.9\% \\
Belly&    8.2\%&   40.7\%&   70.3\%&   81.9\% \\
Breast&    9.8\%&   45.1\%&   74.2\%&   84.5\% \\
Crown&   12.2\%&   59.8\%&  87.7\%&   94.8\% \\
Forehead&   13.2\%&   63.7\%&   91.0\%&   96.0\% \\
Left eye&   11.3\%&   66.3\%&   91.0\%&   95.7\% \\
Left leg&    7.8\%&   33.7\%&   56.6\%&   64.6\% \\
Left wing&    6.7\%&   31.7\%&   56.7\%&   67.8\% \\
Nape&   11.5\%&   54.3\%&   82.9\%&   90.7\% \\
Right eye&   12.5\%&   63.8\%&   88.4\%&   93.8\% \\
Right leg&    7.3\%&   36.2\%&   56.4\%&   64.9\% \\
Right wing&    6.2\%&   33.3\%&   58.6\%&   69.3\% \\
Tail&    8.2\%&   39.6\%&   65.0\%&   74.7\% \\
Throat&   11.8\%&   56.9\%&   87.2\%&   94.5\% \\
\hline
\end{tabular}
\caption{\textbf{Keypoint Localization results.} We use PCK as our evaluation metric.  The prediction is correct if lies within $\alpha \times max(h, w)$ of the annotated keypoint with the corresponding object's dimension being (h, w). We show results on different $\alpha$. No bounding box information is used. }
\label{table:PCK}
\end{table}

\begin{table}[t]
\centering
\begin{tabular}{|c|c|c|}
\hline
& Head & Body \\
\hline
RCNN \citep{eccv14} & 61.42\% & 70.68\% \\
\hline
Ours & \textbf{63.82} \% &  \textbf{89.06}\%\\
\hline
\end{tabular}
\caption{Part localization accuracy in terms of PCP
on the CUB200-2011 bird dataset.}
\label{tab:box_pcp}
\end{table}

For both the localization net and classification net, we fine-tuned from the 16-layer VGG network used in \citep{vgg} pretrained on ImageNet data. Like \citep{fcn}, we change the last two fully connected layer to convolutional layer and make the net fully convolutional. All of our experiments are implemented in the open source Caffe framework \citep{caffe} and we will make our architecture and weights publicly available.

\subsection{Fine-grained classification}
We first present classification results on the standard CUB200-2011 dataset. 
The classification accuracies of our method and other methods are reported in Table \ref{table:finegrainedres}, along with the different annotations required during the training and test phases. Our method with compact bilinear pooling achieves 83.00\% classification accuracy, and fine-tuning part networks improves accuracy to 85.92\%. This is state-of-the-art and improves over other methods supervised by part annotations.


VGG net pretrained on Imagenet alone can achieve 58.34\% classification accuracy and finetuning on our dataset can improve the performance to 71.73\%. By adding pose-normalized representation, our method can achieve 75.04\% accuracy. We also tested the keypoint feature baseline results, i.e. finetune the keypoint localization net using classification loss and the accuracy dropped to 65.10\%. It means it is necessary to split out the keypoint localization network and the fine-grained classification network.

\subsection{Part localization}
We now present results of part localization using the localization net in our method. We use PCK (Percentage of Correctly Localized Keypoints) as our evaluation metric and the results are shown in Table \ref{table:PCK}.  For each annotated instance, the keypoint is said to have been predicted correctly if the corresponding prediction lies within $\alpha \times max(h, w)$ of the annotated keypoint with the corresponding object's dimension being (h, w). For each keypoint, the PCK is measured as the fraction of objects where it was found correctly.

\begin{figure}[t]
\centerline{
\begin{tabular}{cccc}
Groundtruth & Prediction & Groudtruth & Prediction \\
\includegraphics[height=0.25\linewidth]{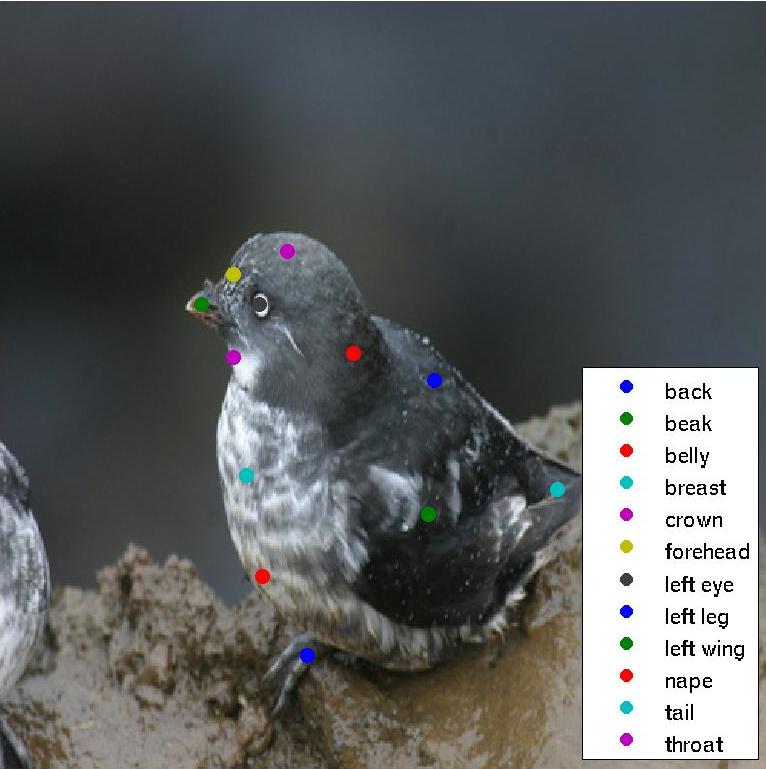} &
\includegraphics[height=0.25\linewidth]{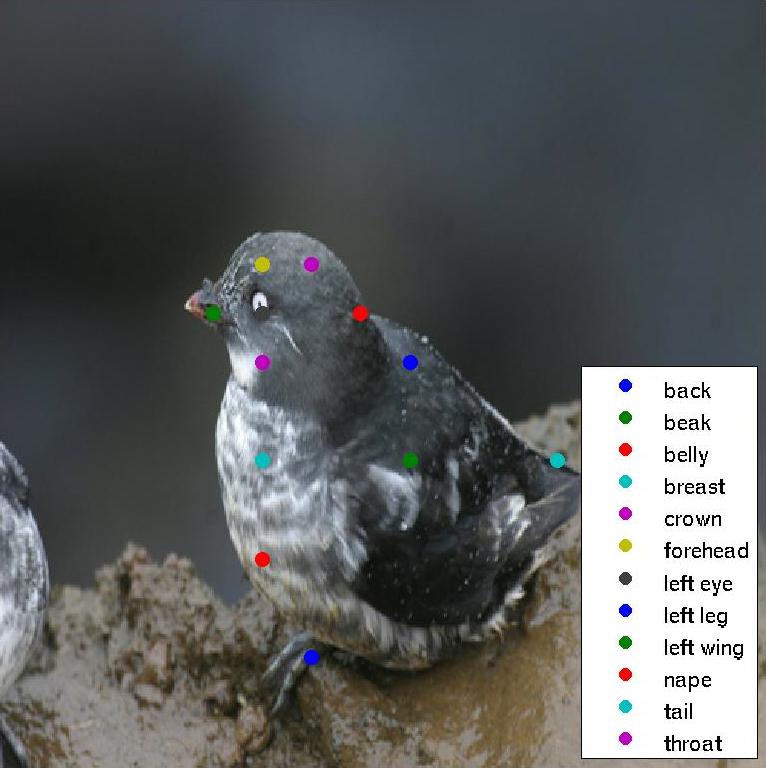} &
\includegraphics[height=0.25\linewidth]{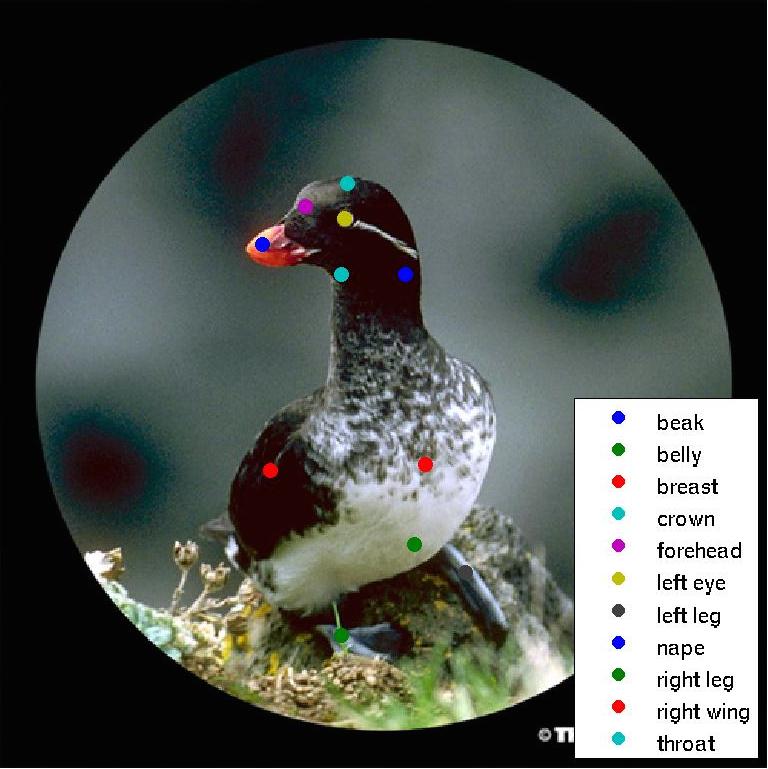} &
\includegraphics[height=0.25\linewidth]{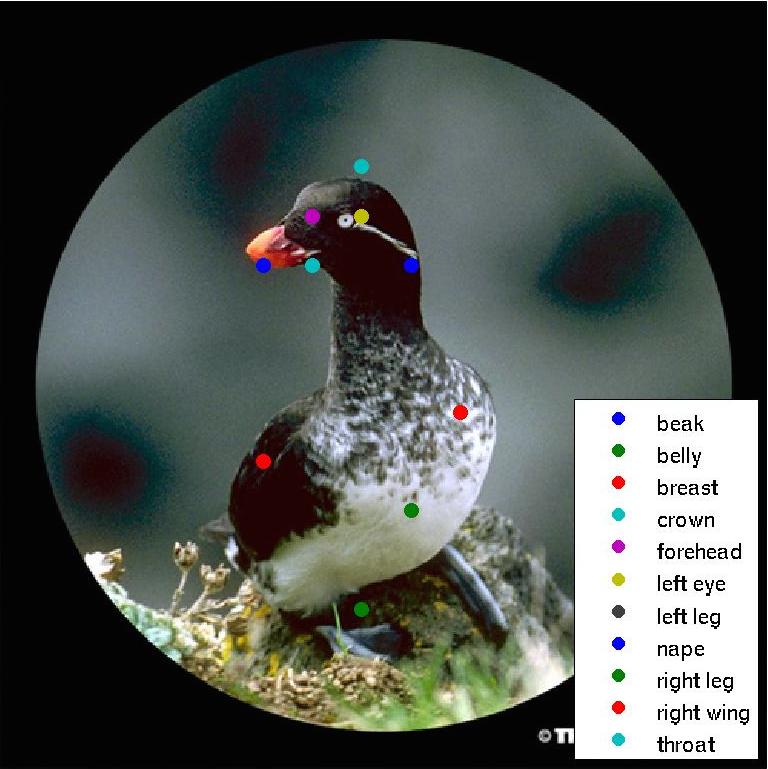}\\
\includegraphics[height=0.25\linewidth]{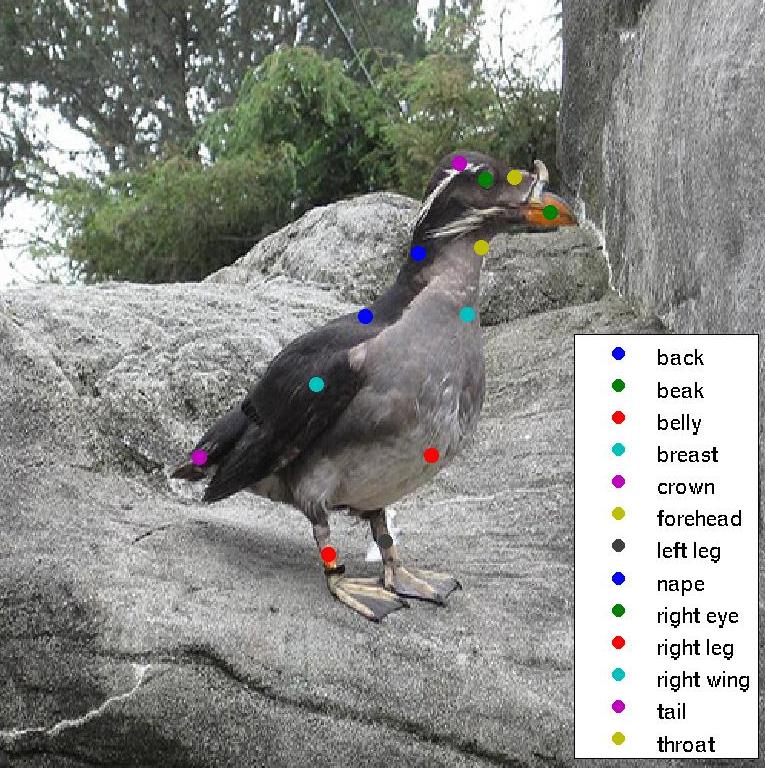} &
\includegraphics[height=0.25\linewidth]{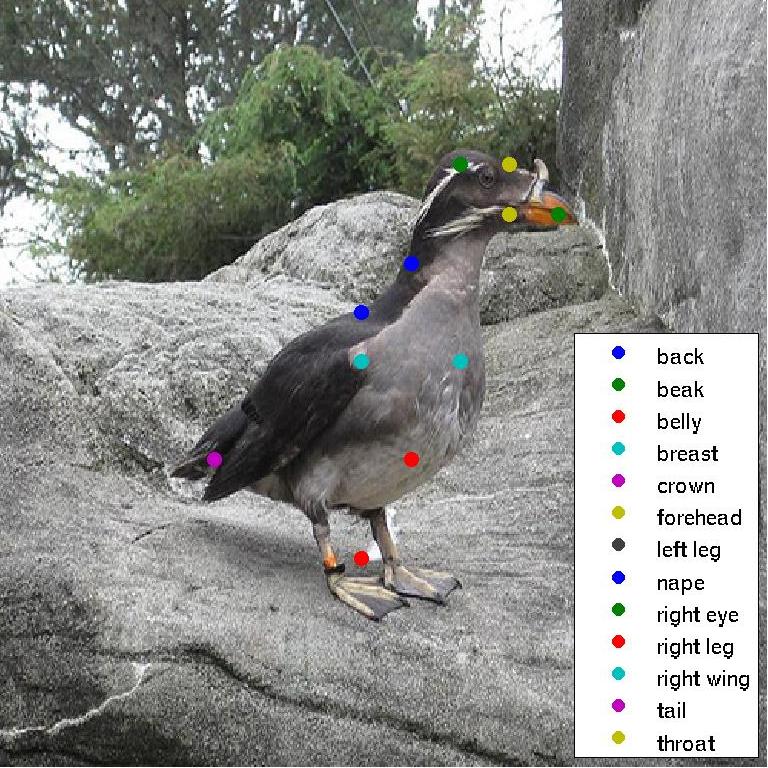} &
\includegraphics[height=0.25\linewidth]{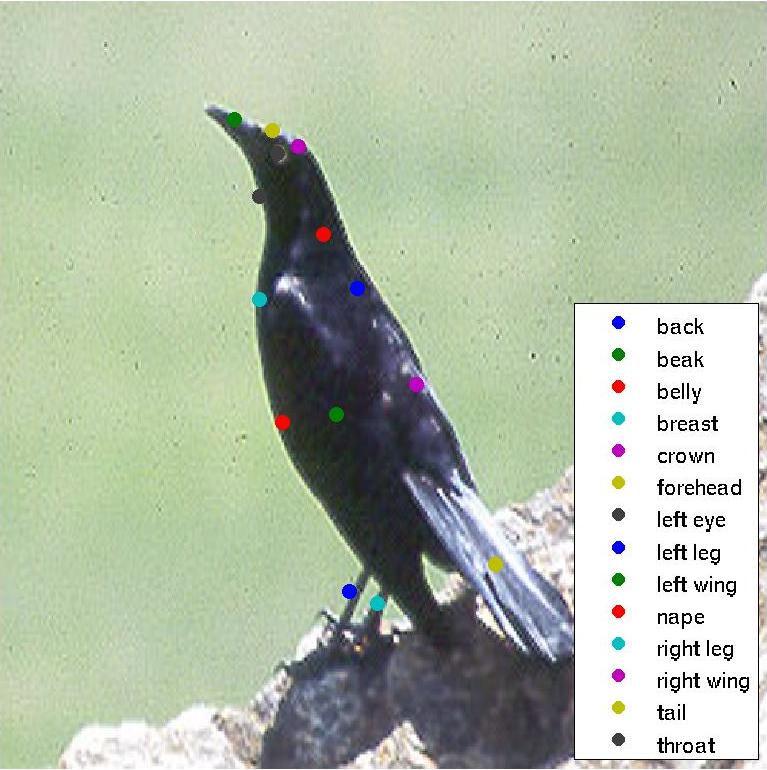} &
\includegraphics[height=0.25\linewidth]{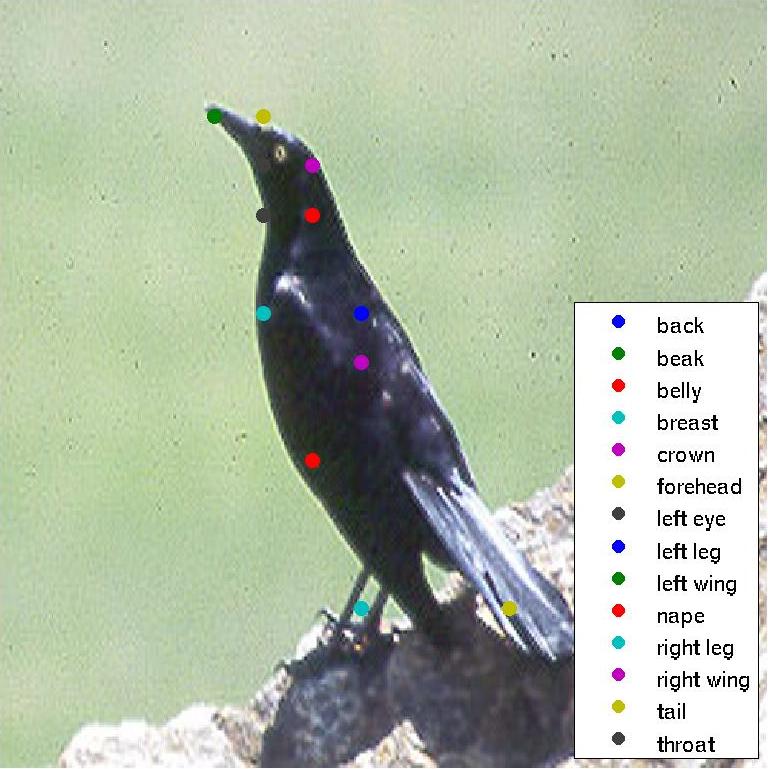}\\
\includegraphics[height=0.25\linewidth]{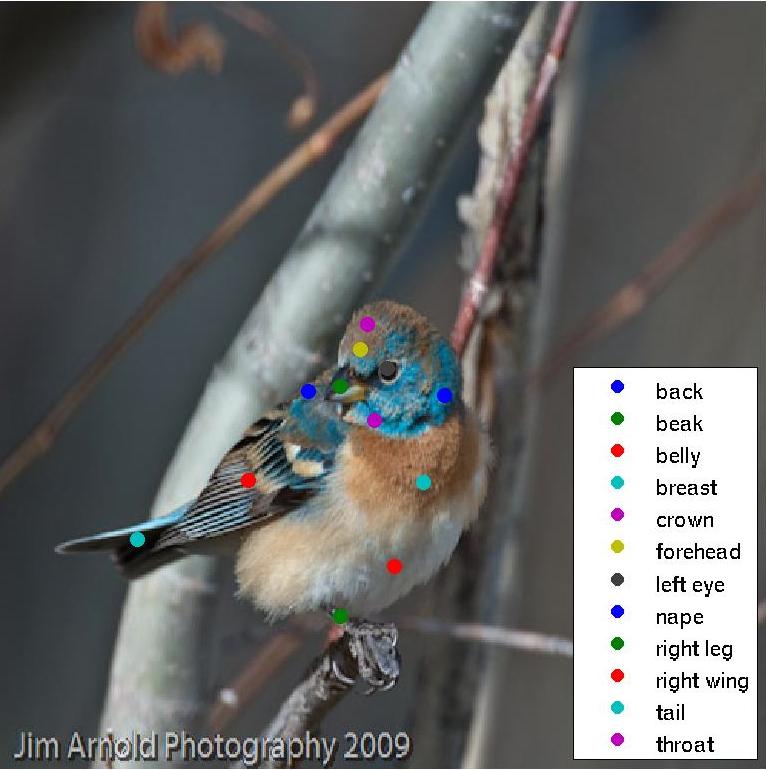} &
\includegraphics[height=0.25\linewidth]{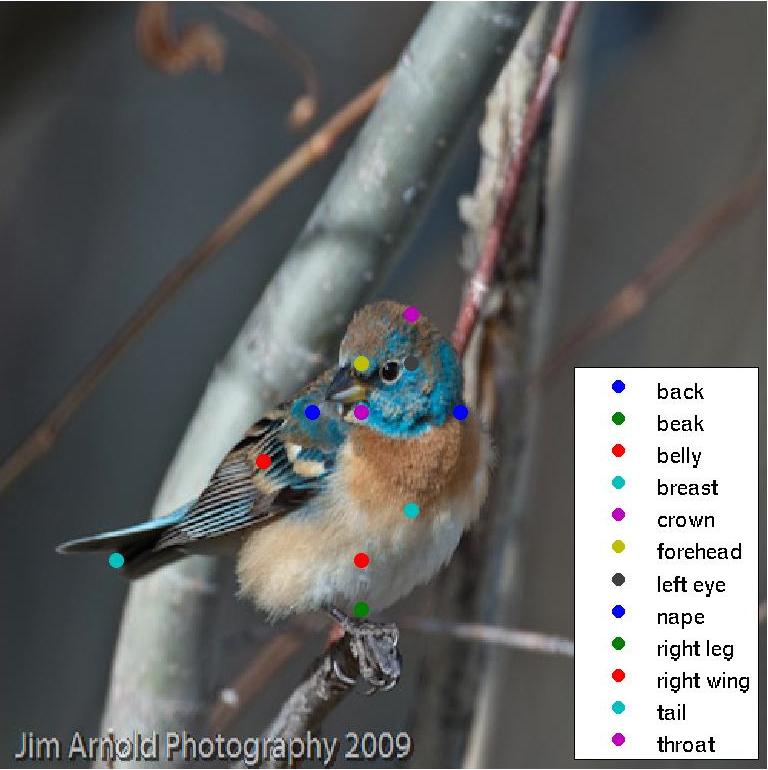} &
\includegraphics[height=0.25\linewidth]{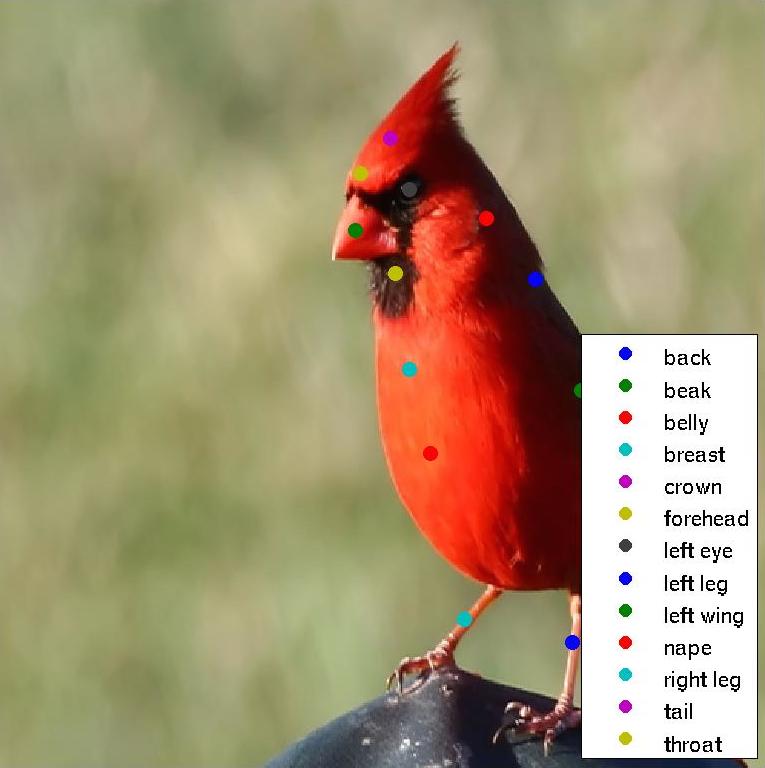} &
\includegraphics[height=0.25\linewidth]{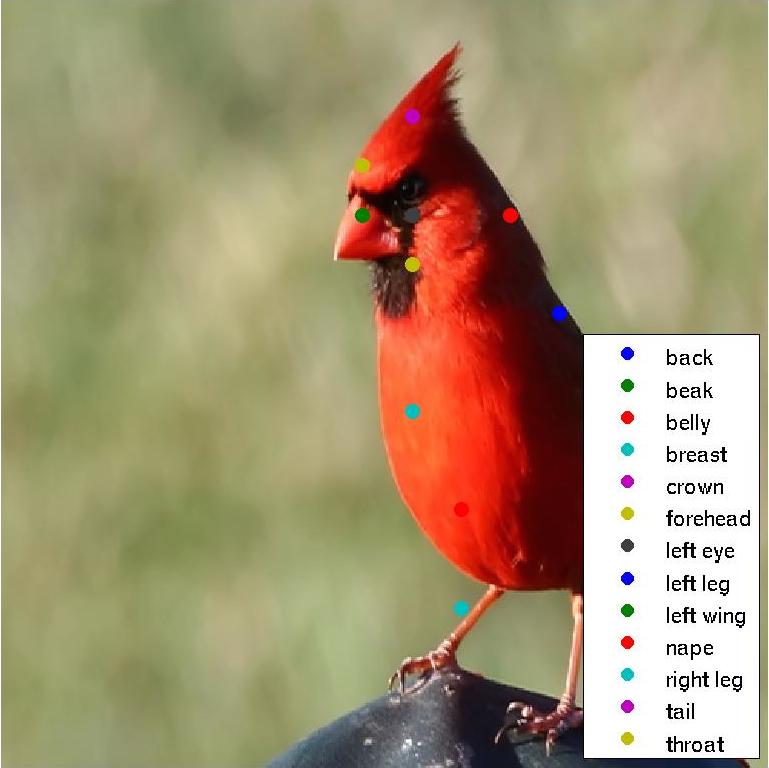}\\
\includegraphics[height=0.25\linewidth]{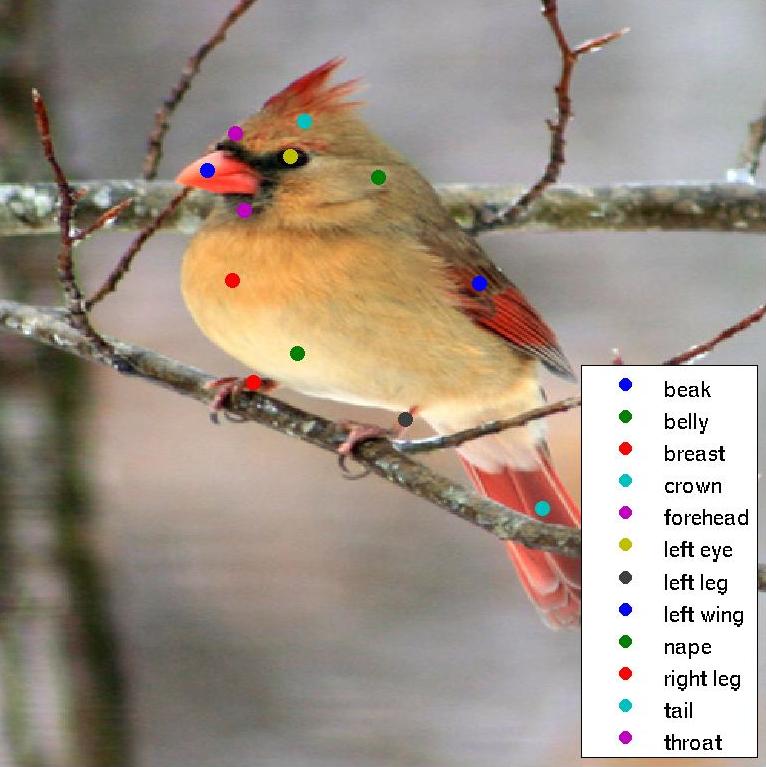} &
\includegraphics[height=0.25\linewidth]{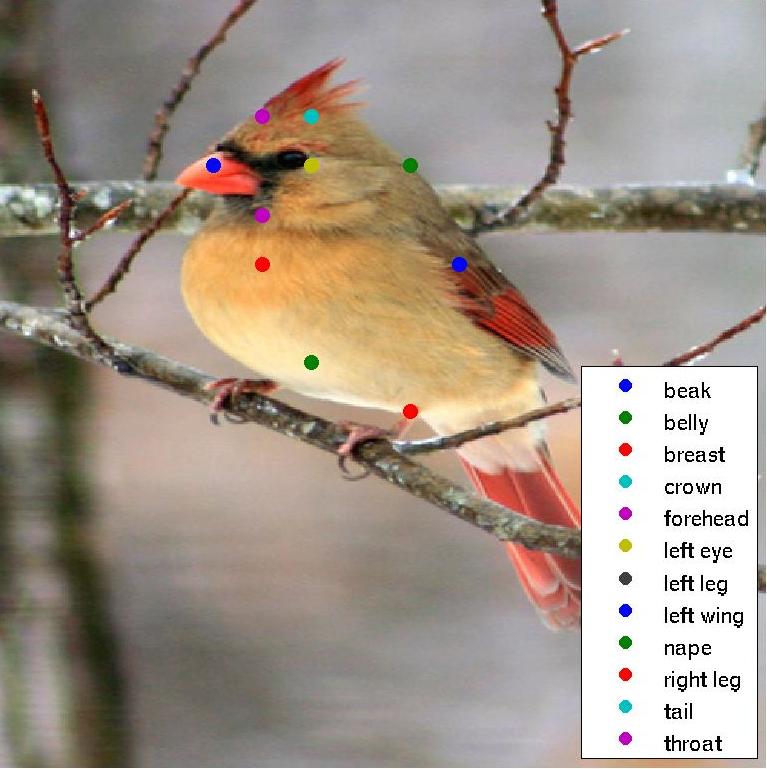} &
\includegraphics[height=0.25\linewidth]{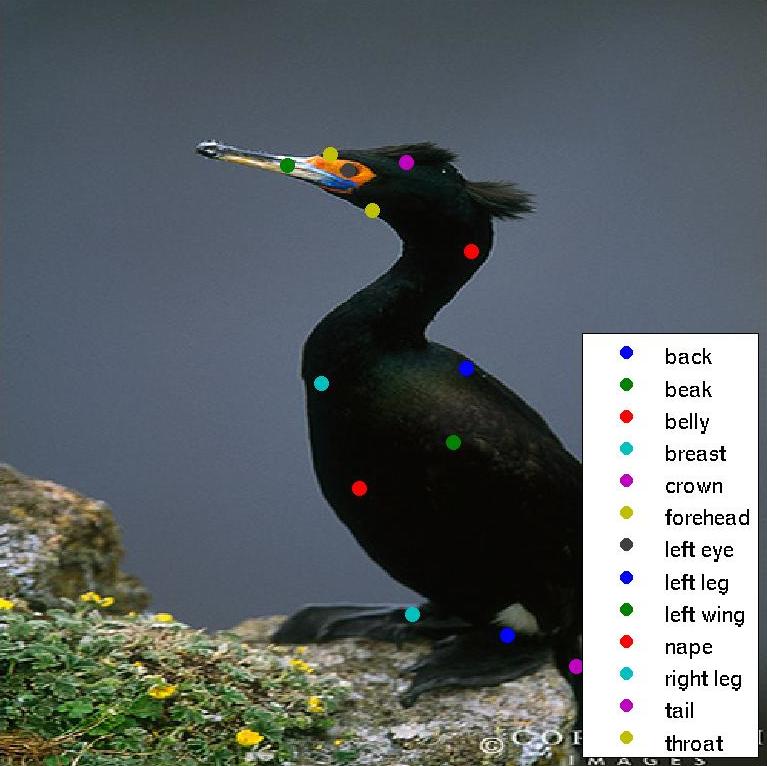} &
\includegraphics[height=0.25\linewidth]{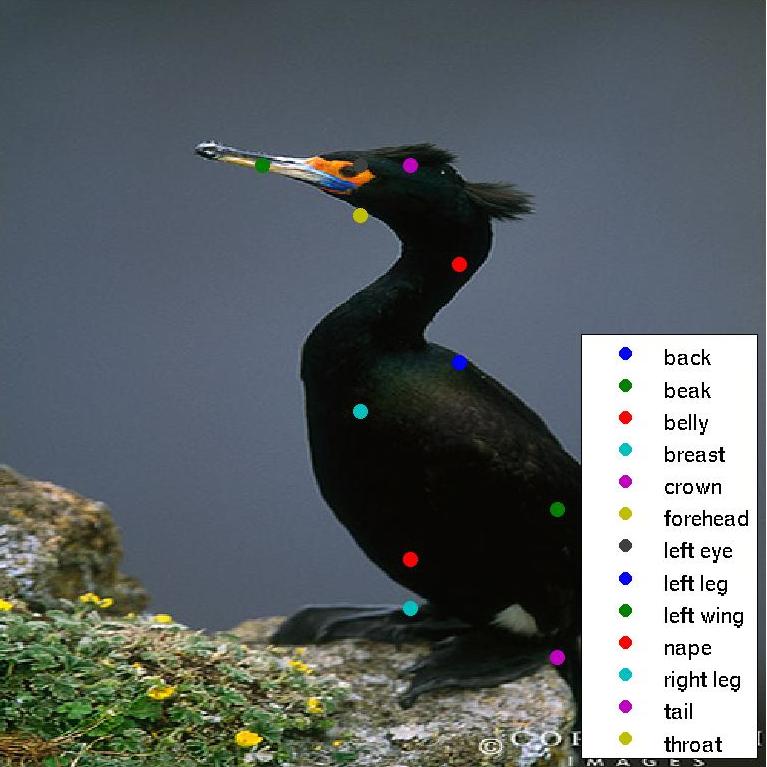}\\
\end{tabular}
}
\caption{Visualizations of keypoint predictions. Ground truth annotations are shown in the left and our prediction results are shown in the right. Each color map to one keypoint. No bounding box information is used during training or test time. The images shown are warped to align the visualizations.}
\label{fig:keypoint_vis}
\end{figure}

As shown in the Table \ref{table:PCK}, although no bounding box information is used, our method can still have strong localization performances even when the requirement $\alpha$ is small.
To our knowledge, these are the first reported PCK numbers on this dataset. We also show part localization visualizations in Figure \ref{fig:keypoint_vis}. We show the ground truth  annotation and our prediction side-by-side. As you can see, almost all of the part predictions lie on the bird body. Some errors are caused by the confusion of left and right, for example, localizing left leg on the position of right leg.
The small localization errors could be corrected by fusing multiple feature layers \citep{fcn, hypercolumn} or multi-scale modeling \citep{TompsonJLB14}.

We also show the localization results on part box prediction in Table \ref{tab:box_pcp} in terms of PCP (Percentage of correctly localized parts). The prediction is correct if the overlap between it and grountruth part box is over 0.5. We compared with \citep{eccv14} and we outperform the RCNN detection. 

\section{Conclusion}
Fine-grained classification can be accomplished using part-aligned representations that reveal subtle distinguishing features of different classes. Learned part localization and feature transformation networks can discount the nuisance variables of pose and viewpoint.

We have presented an approach to classification that embodies fine-grained approaches to part/keypoint localization, pose-normalized descriptor learning, and category learning.  Our model is fine-grained in its spatial localization ability, and in its ability to distinguish closely related classes in a fine-grained recognition task.  A fully convolutional network can localize instance keypoints using a pixel-level map, and deep pose-normalized descriptors can make distinctions across fine-grained category labels learned in a coordinate space that is defined using the estimated keypoints.   

We unify these steps in an end-to-end trainable network that estimates part locations, pose-normalized descriptor representations, and fine-grained classifier weights using a joint loss on keypoint predictions and the classification task.

A careful comparison of strong keypoint supervision and weak class label supervision is valuable future work to guide further improvements to fine-grained recognition. While strong supervision aids with correspondence, weak supervision may learn from more data. This comparison could take the form of a strongly supervised spatial transformer network or an unsupervised variant of our network incorporating part discovery as in \citep{simon2015neural}. With end-to-end multi-task learning it should be possible to unify these approaches in semi-supervised models for fine-grained recognition.


{
\small
\bibliographystyle{iclr2016_conference}
\bibliography{reference}
}
\end{document}